\documentclass{llncs}
\usepackage{graphicx}

\begin{document}           
\title{Applying MAPP Algorithm for Cooperative Path Finding in Urban Environments}
\author{Anton Andreychuk\inst{1, 2} \and Konstantin Yakovlev\inst{1, 3}}
\institute{Federal Research Center ``Computer Science and Control'' of Russian Academy of Sciences, Moscow, Russia \and People's Friendship University of Russia (RUDN University), Moscow, Russia \and National Research University Higher School of Economics, Moscow, Russia\\ \email{andreychuk@mail.com, yakovlev@isa.ru}}
\maketitle
\begin{abstract}
The paper considers the problem of planning a set of non-conflict trajectories for the coalition of intelligent agents (mobile robots). Two divergent approaches, e.g. centralized and decentralized, are surveyed and analyzed. Decentralized planner -- MAPP is described and applied to the task of finding trajectories for dozens UAVs performing nap-of-the-earth flight in urban environments. Results of the experimental studies provide an opportunity to claim that MAPP is a highly efficient planner for solving considered types of tasks.
\end{abstract}

\section{Introduction}
Planning a set of non-conflict trajectories for a coalition of intelligent agents is a well known problem with the applications in logistics, delivery, transport systems etc. Unlike path-finding for a single agent (robot, unmanned vehicle etc.) for which numerous computationally-effective algorithms exists \cite{koenig2002}\cite{magid2006}\cite{yakovlev2015}\cite{harabor2016} the multi-agent path finding problem  lacks an efficient general solution and belongs to the PSPACE hard problems \cite{hopcroft1984}. Therefore the development of computationally efficient algorithms for cooperative path finding is actual yet challenging task.

Commonly in robotics and Artificial Intelligence path finding is considered to be a task of finding the shortest path on a graph which models agent's environment. Vertices of such graph correspond to the locations of the workspace the agent can occupy and edges correspond to the transitions between them, e.g. to the elementary trajectories the agent can follow in an automated fashion (segments of straight lines, curves of the predefined shape etc.). In 2D path finding, which is under consideration in this work, the most often used graph models are: visibility graphs \cite{lozano1979}, Voronoi diagrams \cite{bhat2008}, navigation meshes \cite{kallmann2010}, regular grids \cite{yap2002}. The latter are particularly widespread in robotics \cite{elfes1989} as they can be constructed (and updated) on-the-fly when processing the sensors’ input.

Formally, grid is a weighted undirected graph. Each element (vertex) of the grid corresponds to some area of regular shape (square in considered case) on the plane, which can be either traversable or untraversable for the agent (mobile robot, UAV etc.). In this work we assume that grid vertices are placed in the centers of the squares the agent’s workspace is tessellated into. The edges are pairs of cardinally adjacent traversable vertices. Graphically, grid can be depicted as a table composed of black and white cells (see Fig. 1). Path on a grid is a sequence of grid edges between two defined vertices, e.g. \textit{start} and \textit{goal}. When augmenting path with time we end up with the trajectory. Typically, in grid based path finding time is discretized into timesteps 1, 2, 3, \ldots, and agent is allowed to traverse a single grid edge per one timestep or stand still occupying some vertex (grid cell). We follow this approach in the paper as well and state the cooperative path finding as defined in Section 2. We then overview the existing methods to solve the problem (Section 3) and describe one of the prominent algorithm, e.g. MAPP, tailored to solve it (Section 4). We experimentally evaluate MAPP in solving navigation tasks for the coalitions of the UAVs (up to 100) performing nap-of-the-earth flight in urban environments and provide the results of this evaluation in Section 5. Section 6 concludes.

\section{Formal Statement}
Consider a square grid which is a set of cells, $L$, composed of two mutually disjoint subsets, e.g. blocked (untraversable) cells, $L^-$, and unblocked (traversable) cells, $L^+:L=L^+\cup L^-=\{l\}$. \textit{U} -- is a non-empty set of agents navigating on the grid. Each agent is allowed to perform cardinal moves, e.g. can go left, right, up or down (from one traversable cell to the other). Path for a single agent \textit{u} from given start and goal cells, e.g. $s_u$ and $g_u$, is a sequence of cardinally adjacent traversable cells: $\pi(u)=(l_0^u,l_1^u,\ldots,l^u_{k_u}), l^u_0=s_u, l^u_{k_u}=g_u, \forall l^u_i\in\pi(u):l^u_i\in L^+, l^u_i\in c_{adj}(l^u_{i-1})$. Here $c_{adj}(l)$ denotes a set of cardinally adjacent cells for the cell $l$. At the initial moment of time the agents are located at the corresponding start cells. During a timestep each agent can move from the cell $l^u_i$ to the cell $l^u_{i+1}$ or stay at place. The set of moments of time (timesteps) at which agent \textit{u} occupies the cell $l^u_i$ is denoted as $time(l^u_i)$. The cooperative path finding task is to build a set of non-conflicting trajectories, i.e. such paths that the following conditions are met: $\forall u,v\in U,\forall l^u_i=l^v_j: time(l^u_i)\cap time(l^v_j)=\emptyset, i\in[0,k_u],j\in[0,k_v]$.

\vspace*{-\baselineskip}
\begin{figure}
	\centering
		\includegraphics[scale=0.33]{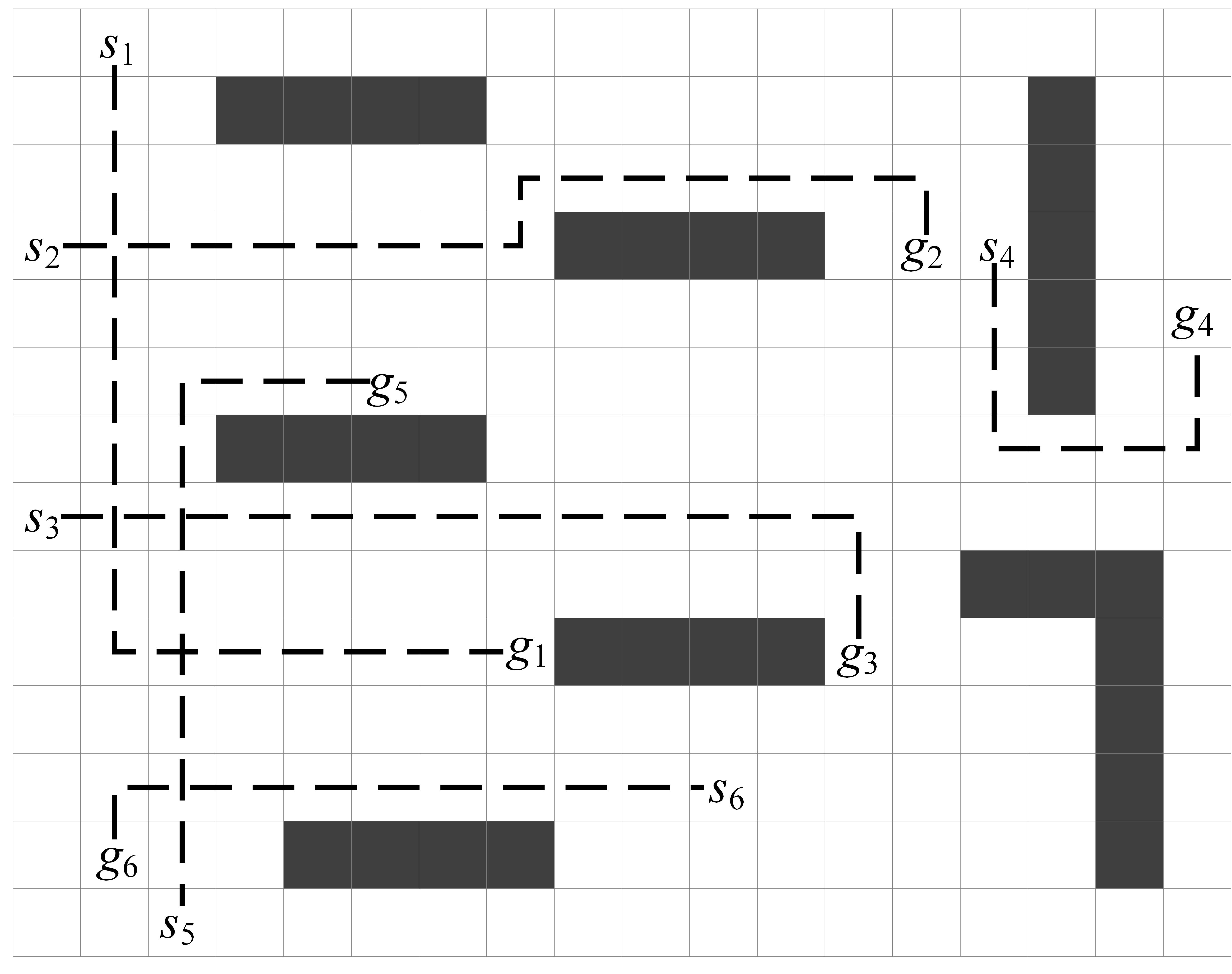}
		\caption{Graphical representation of the grid based cooperative path finding task. Light cells correspond to traversable vertices, dark cells - to untraversable ones. Dotted lines show agents' paths.}
	\label{fig1}
\end{figure}

\section{Methods and Algorithms}
Cooperative (multi-agent) path finding algorithms can be divided into two groups: centralized (coupled) and decentralized (decoupled) planners. Centralized planners search trajectories for all agents simultaneously, taking into account the position of every agent at each moment of time. Such algorithms are theoretically complete, e.g. guarantee finding a solution if it exists, and can be optimal. Decentralized planners typically build on top of assigning priorities to the agents, finding trajectories one-by-one according to these priorities and eliminating conflicts. They are much faster than the centralized ones but do not guarantee optimality (and even completeness) in general.

One of the well known optimal algorithm implementing the centralized approach is the OD+ID algorithm described in \cite{standley2010}. OD+ID builds on top of A* \cite{hart1968}. The distinctive feature of OD+ID  is the use of the procedures that allow speeding up the search and reducing the search space. The first procedure, Independence Detection (ID), divides the original task into several subtasks in such a way that the optimality of the sought solution is not violated. The second - Operator Decomposition (OD), is used to reduce the search space. Instead of moving all agents at the same time, agents move in turn, thereby reducing the number of different states from $M^n$ to $M$$*$$n$, where $M$ is the number of agents and $n$ is the number of possible moves for each agent (4 in considered case).

One of the most advanced centralized planners up-to-date is the CBS (Conflict Based Search) algorithm \cite{sharon2015} and its variations like ICBS\cite{boyarski2015}. It's a two level search algorithm. At the low level, it searches a path for a single agent and every time it finds a conflict, it creates alternative solutions. At the high level, the algorithm builds a tree of alternative solutions until all conflicts are eliminated. Due to the fact that among the alternative solutions the best is always chosen, the optimality of the overall solution is guaranteed. Two level decomposition leads to reducing the search space and increasing the efficiency of the algorithm as compared to other centralized planners.

Another well-known centralized and optimal planner is M*\cite{wagner2011}. It relies on the technique called subdimensional expansion, which is used for dynamical generation of low dimensional search spaces embedded in the full configuration space. It helps to reduce the search space in some cases. But if the agents cannot be divided into independent subgroups (and this can happen quite often in practice), the introduced technique doesn't work well and M* spends too much time to find a solution.

In general, optimal algorithms for cooperative path planning that implement centralized approach are very resource-intensive and do not scale well to large problems when the number of agents is high (dozens or hundreds) and the search space consists of tens and even hundreds of thousands of states.

There exist a group of centralized algorithms that trade off the optimality for computational efficiency.

One of such algorithms is OA (Optimal Anytime) \cite{standley2011}, proposed by the authors of OD+ID. The principle of its work is similar to the OD+ID algorithm. The difference is that when conflicts are detected, agents do not try to build joint optimal trajectories. They join the group, but build their own paths, considering other agents only as moving obstacles. This approach works faster, but does not  guarantee the optimality of the solution any more.

There also exist suboptimal variants of CBS, such as ECBS\cite{barer2014}, for example.

The second large class of the cooperative path finding algorithms is a class of the algorithms utilizing a decentralized approach. These algorithms are based (in one or another way) on splitting the initial problem into several simpler problems which can be viewed as a form of prioritized planning first proposed in \cite{erdmann1987}. The algorithms implementing a decentralized approach are computationally efficient, but do not guarantee finding the optimal solutions and are not even complete in general.

One of the first and most applicable algorithms in the gaming industry, using a decentralized approach, is the LRA* (Local Repair A*) algorithm proposed in \cite{zelin1992}. The algorithm works in two stages. At the first stage, each agent finds an individual path using A* algorithm, ignoring all other agents. At the second stage, agents move along the constructed trajectories until a conflict arises. Whenever an agent cannot go any further, because another agent is blocking its path, the former re-plans its path. How-ever, due to the fact that each agent independently reshapes its path, this leads to the emergence of new conflicts and as a result, there may arise deadlocks (cycle conflicts) that cannot be resolved.

WHCA* (Windowed Hierarchical Cooperative A*) algorithm described in \cite{silver2005} is one of the algorithms eliminating the shortcomings of LRA* algorithm. Like LRA*, WHCA * algorithm works in two stages and uses the A* algorithm to find individual paths. The improvement of the algorithm lies in the fact that the search for the path for each agent occurs in the ``window'' - an abstract area of the environment model (grid) having the size $w$$*$$w$, which is gradually shifts towards the goal. The algorithm uses a 3-dimensional space-time table, in which the positions of all agents are stored at each timestep. Using such a table can reduce the number of conflicts that arise between agents, thereby increasing the speed of the algorithm. However, WHCA* algorithm is also not complete, i.e. in some cases it does not find a solution, even though it exists. Another disadvantage of the algorithm is its parametrization, i.e. the dependence on the parameter $w$.

One of the last modifications of WHCA* -- Conflict Oriented WHCA* (CO-WHCA*) \cite{bnaya2014} combines ideas from both WHCA* and LRA*. It uses a technique that focuses on areas with potential conflicts between agents. The experiments carried out by the authors of the algorithm showed that it finds more solutions than WHCA* and LRA*, but is still not complete.

There also exist decentralized (decoupled) planners that guarantee completeness under well-defined constraints \cite{wang2009}, \cite{cap2015}. These algorithms are of particular interest to our study as they scale well to large problems, find solutions fast while the restrictions they impose on class of tasks they guarantee to solve are likely to be held in the scenarios we are looking at (navigating dozens of UAVs in urban environments). We choose MAPP algorithm for further consideration as it is originally tailored to solve multi-agent path finding tasks on 4-connected grid-worlds (as in considered case).

\section{Algorithm MAPP}
MAPP algorithm is a prominent cooperative path planner \cite{wang2009}, \cite{wang2011} that utilizes decentralized approach, but in contrast to other decentralized algorithms mentioned above, is complete for a well-defined class of problems called Slidable.

An agent $u\in U$ is Slidable if a path $\pi(u)=(l^u_0, l^u_1, \ldots, l^u_{k_u})$ exists and the next three conditions are met:

\begin{enumerate}
	\item {Alternative connectivity. For each three consecutive locations $l^u_{i-1}, l^u_i, l^u_{i+1}$ on $\pi(u)$ there exists an alternative path $\Omega^u_i$ between $l^u_{i-1}$ and $l^u_{i+1}$ that does not include $l^u_i, i=\overline{(1, k_u-1)}$ (see Fig.2).}
	\item{Initial blank. At the initial timestep, when all agents are at $l_0^u$, $l_1^u$ is blank, i.e. unoccupied, for each agent $u\in U$.}
	\item {Goal isolation. Neither $\pi$ nor $\Omega$-paths include goal locations of other agents. More formally:
	\begin{enumerate}
	    {
		\item{$\forall v\in U$\textbackslash$\{u\}:g_u\notin\pi(v)$}
		
		\item{$\forall v\in U,\forall i=\overline{(1, k_v-1)}: g_u\notin\Omega^v_i$}
		}
    \end{enumerate}
	}
\end{enumerate}

Problem belongs to the class Slidable if all agents $u \in U$ are Slidable.

\begin{figure}
	\centering
		\includegraphics{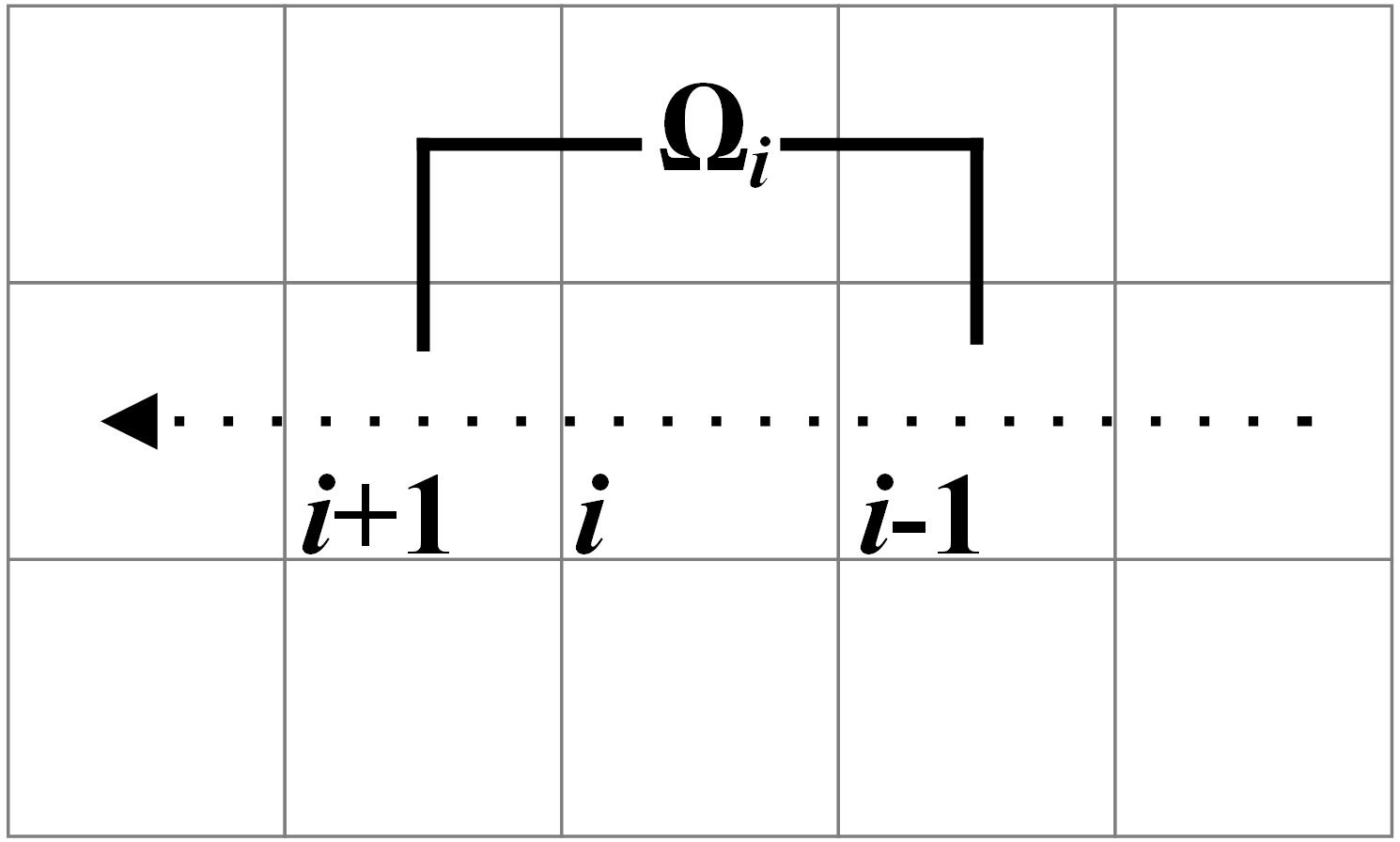}
		\caption{An example of an alternative path $\Omega_i$.}
	\label{fig2}
\end{figure}

At the first stage MAPP finds individual Slidable paths with the enhanced A* algorithm. The modifications of the original A* \cite{hart1968} are the following. First MAPP's A* checks for a blank location in the first step. Second it always avoids passing through other goals. Third it ensures that alternate connectivity condition is always met, e.g. when expanding a node $x'$, its neighbor $x''$ is added to the set of potential candidates (so-called OPEN list) only if there exists an alternate path between $x''$ and $x$, the parent of $x'$. Thus, MAPP guarantees that each node in path $\pi$ (if it is found) can be excluded and replaced by the alternative path $\Omega$ (see Fig. 2).

After the first stage, all agents, for which Slidable paths are found, move step by step to their goals in the predefined order (from high-priority agents to low priority ones). To ensure that lower priority agent do not interfere with any of the higher priority agents a concept of the private zone is used. The private zone includes the current location of the agent and the previous position in case an agent is moving along the computed path. Disallowing the lower priority agents to move through the private zones of higher priority agents one can guarantee that at least one agent with the highest priority will move closer to its goal at each step. The procedure for progression step is described in Algorithm 1. The figure is taken from \cite{wang2011}.

\begin{figure}
	\centering
		\includegraphics[scale=0.45]{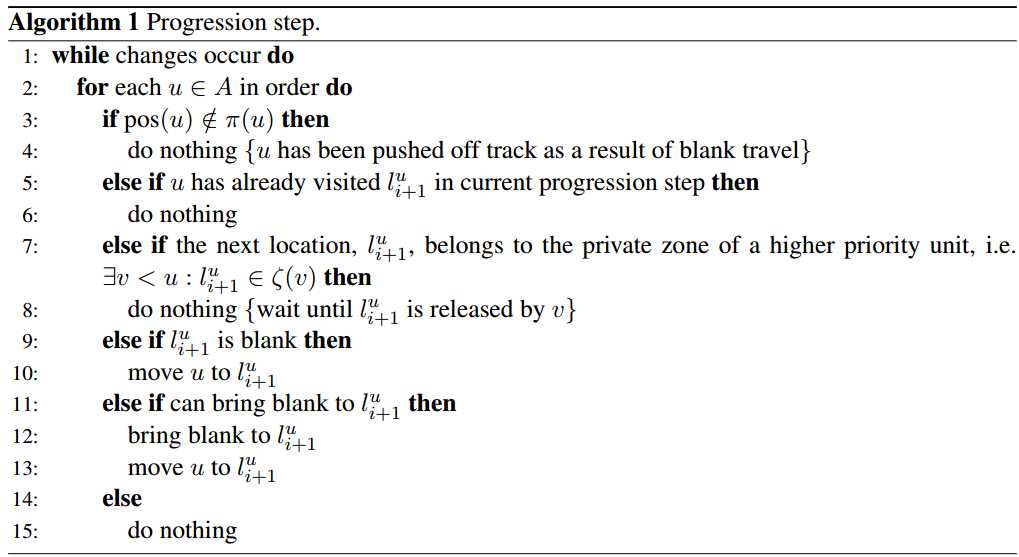}
		\caption{Pseudocode of the progression step (from \cite{wang2011}).}
	\label{fig3}
\end{figure}

Lines 3-15 in Algorithm 1 show the processing of current agent $u$. If $u$ has been pushed off its precomputed path, then it stays in its current location and does nothing (lines 3-4). Lines 5 and 6 cover the situation when agent $u$ has been pushed around (via blank travel) by higher-priority agents back to a location on $\pi(u)$ already visited in the current progression step. If $u$ is on its path but the next location $l_{i+1}^u$ is currently blocked by an agent $v$ with higher priority, then it just waits (lines 7-8). Otherwise, if the next location $l_{i+1}^u$ is available, $u$ moves there (lines 9-10). Finally, if $l_{i+1}^u$  is occupied by a lower priority agent, then it brings blank to $l_{i+1}^u$ and moves agent $u$ there (lines 11-13). When $u$ moves to $l_{i+1}^u$, algorithm checks if $l_{i+1}^u$ is the goal location of $u$. If this is the case, then $u$ is marked as solved by removing it from $A$ and adding it to $S$, the set of solved agents.

MAPP algorithms guarantees that all Slidable agents will reach their destinations. The proof of this statement is based on the fact that at least one agent, the one with the highest priority, gets closer to its goal at each step. After a finite number of steps this agent will reach the goal, the highest priority will go to some other agent and so on. Thus at the end all agents will achieve their goals without colliding with other agents. Non-Slidable agents will simply stay at their start locations.

\section{Experimental Evaluation}
We were interested in scenarios of finding non-conflict trajectories for numerous (dozens of) UAVs performing nap-of-the-earth flight in urban areas. The considered scenario assumes that all UAVs fly at the same altitude and that allows us to consider the problem as a search for non-conflict trajectories on the plane. It was shown earlier in \cite{sharon2015} that existing optimal algorithms based on the centralized approach, are poorly applicable for the problems involving large number ($>$85) of agents acting in similar environments, as they need too much time (more than 5 minutes) to solve a task. Therefore, the main goal was to examine whether MAPP algorithm will succeed. We fixed the number of agents (UAVs) to be 100.

To run experiments MAPP algorithm was implemented from scratch in C++. The experiments were conducted using Windows-7 PC with Intel QuadCore CPU (@ 2.5GHz) and 2 Gb of RAM.

The algorithm was tested on a collection of 100 urban maps. Each map represents a $~$1.81 km$^2$ fragment of the real city environment as it was extracted from the OpenStreetMaps geo-spatial database\footnote{http://wiki.openstreetmap.org/wiki/Database}. An example is depicted on Fig. 4. Each map was discretized to a 501x501 grid. All 100 maps were chosen in such a way that the number of blocked cells was  20-25\% on each of them.

Path finding instances (sets of start/goal locations) of two types were generated (type-1 and type-2 tasks). For type-1 instances start and goal locations were chosen randomly at the opposite borders of the map (see Fig. 4-left), resulting in sparse distribution of the start/goals. For the type-2 instances compact start and goal zones on the opposite borders of the map were chosen and all start/goals were confined to these zones (see Fig. 4-right). The size of each zone was 50x50 cells. For each map two instances of each type were generated. Thus the testbed consisted of 400 tasks in total.

\begin{figure}
	\centering
		\includegraphics[scale=0.7]{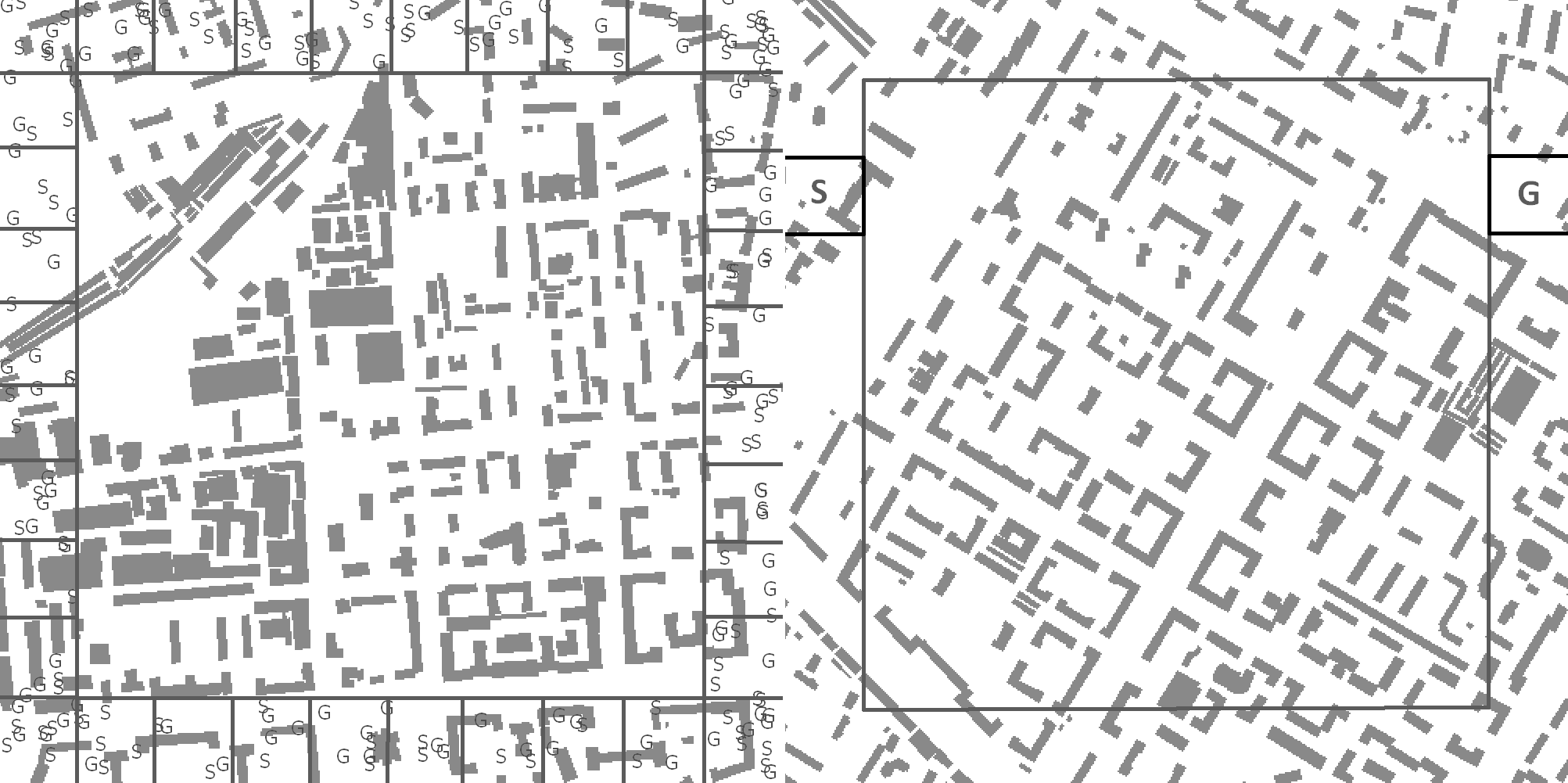}
		\caption{Fragments of the city maps that were used for experimental evaluation.}
	\label{fig4}
\end{figure}

To assess the efficiency of the algorithm, the following indicators were tracked:
\begin{itemize}
\item{Time - algorithm's runtime. Measured in seconds.}
\item{Memory - the maximum number of nodes that the algorithm stores in memory during the path planning for an individual agent.}
\item{Path length - the number of transitions between the grid cells plus the waiting time, averaged for all solved agents of the instance.}
\item{Conflicts - the number of times high-priority agent is blocked by the low-priority one.}
\item{Conflict agents - number of the agents that have at least one conflict.}
\item{Success - the percentage of agents for which Slidable trajectories were successfully found.}
\end{itemize}

Table 1 presents the averaged results of the experiment. First row in the table contains results for the type-1 tasks, second row for the type-2 tasks.

As one can see, MAPP successfully solved all the type-1 instances (when the agents were disseminated sparsely close to the map borders), although for the type-2 tasks the success rate was 99.97\%. In some cases (6, to be precise) MAPP was not able to find Slideble trajectory for one agent and this was enough to deem the task failed.

Obviously, type-2 tasks are more complex in terms of conflict resolution. All agents move from one constrained zone to another and therefore their trajectories often overlap. Due to this, there are almost 6 times more conflicts (compared to type-1 instances) and more than 85\% of agents have at least one conflict. However, this almost does not affect the runtime, since most of the time is spent on building the individual trajectories.

\begin{table}[t]
	\caption{ Averaged results of experimental evaluation of MAPP.}
	\centering
		\begin{tabular}{| c | c | c | c | c | c |}
			\hline
			\textbf{\#} & \textbf{Time(s)} & \textbf{Memory(nodes)} & \textbf{Pathlength} & \textbf{Conflicts} & \textbf{Success} \\
			\hline
			1	& 32.606	& 84955	& 540.457	& 33.015 & 100\% \\
			\hline
			2	& 31.398	& 59362	& 540.559	& 192.31 & 99.97\% \\
			\hline
		\end{tabular}
	\label{tab1}
\end{table}

Solving type-2 tasks consumes notably less memory. This can be explained in the following manner. Start and goal locations for type-1 tasks were located in different areas of maps. Thus for almost all tasks, chances are at least one of the agents needs to build complex trajectory, enveloping a large number of obstacles, to accomplish its individual task. Building such complex paths necessarily leads to processing (and storing in memory) a large number of search nodes. On the opposite, individual trajectories for type-2 tasks were approximately the same, due to the fact start and goals for all agents were located in the same compact regions. Therefore, for some tasks, when there were few obstacles between the start and goal zones, the  memory consumption was small and this influenced the final averaged value.

Obtained results show that MAPP algorithm can be successfully applied to solving both types of tasks. The performance indicators of the algorithm are high showing its computational efficiency for solving considered class of cooperative path finding scenarios.

\section{Conclusion}
In this paper the problem of cooperative path finding for a set of homogeneous agents was considered. An analysis of existing methods for solving this problem was presented. MAPP algorithm was described, implemented and evaluated experimentally. The results of the conducted experiments confirmed its ability to efficiently solve cooperative path finding problems for large number (one hundred) of UAVs performing nap-of-the-earth flight in urban environments.
\newline
\newline
\textbf{Acknowledgements.} This work was supported by the Russian Science Foundation (Project No. 16-11-00048).

\end{document}